\documentclass[letterpaper, 10 pt, conference]{ieeeconf}  % Comment this line out if you need a4paper

\IEEEoverridecommandlockouts                              % This command is only needed if 
\usepackage[utf8]{inputenc}
\usepackage[T1]{fontenc}

\usepackage{graphicx}
\usepackage{subcaption}
\usepackage{stfloats}

\usepackage{booktabs}
\usepackage{makecell}
\usepackage{adjustbox}

\usepackage{amsmath,amssymb,amsfonts}
\usepackage{algorithmic}

\usepackage{xcolor}
\usepackage{soul}

\usepackage{cite}

\PassOptionsToPackage{hyphens}{url}  % ensure url allows hyphenation
\usepackage{xurl}                    % loads url internally, with hyphens
\usepackage{hyperref}                % must come after xurl

\usepackage{wrapfig}
\usepackage{comment}
\usepackage{acronym}
\usepackage{textcomp}
\newcolumntype{C}[1]{>{\centering\arraybackslash}m{#1}}

\title{\LARGE \bf
Digital Twins \& ZeroConf AI: Structuring Automated Intelligent Pipelines for Industrial Applications
}

\author{Marco Picone, Fabio Turazza, Matteo Martinelli, Marco Mamei% <-this % stops a space
\thanks{Authors are with the Department of Sciences and Methods for Engineering (DISMI) of the University of Modena and Reggio Emilia, Via G. Amendola 2, Pad. Morselli, Reggio Emilia, Italy. Mail addresses: {\tt\small \{name.surname@unimore.it\}}}%}
}

\begin{document}

\maketitle
\thispagestyle{empty}
\pagestyle{empty}

%%%%%%%%%%%%%%%%%%%%%%%%%%%%%%%%%%%%%%%%%%%%%%%%%%%%%%%%%%%%%%%%%%%%%%%%%%%%%%%%
\begin{abstract}
The increasing complexity of Cyber-Physical Systems (CPS), particularly in the industrial domain, has amplified the challenges associated with the effective integration of Artificial Intelligence (AI) and Machine Learning (ML) techniques. Fragmentation across IoT and IIoT technologies, manifested through diverse communication protocols, data formats and device capabilities, creates a substantial gap between low-level physical layers and high-level intelligent functionalities. Recently, Digital Twin (DT) technology has emerged as a promising solution, offering structured, interoperable and semantically rich digital representations of physical assets. Current approaches are often siloed and tightly coupled, limiting scalability and reuse of AI functionalities. This work proposes a modular and interoperable solution that enables seamless AI pipeline integration into CPS by minimizing configuration and decoupling the roles of DTs and AI components. We introduce the concept of Zero Configuration (ZeroConf) AI pipelines, where DTs orchestrate data management and intelligent augmentation. The approach is demonstrated in a MicroFactory scenario, showing support for concurrent ML models and dynamic data processing, effectively accelerating the deployment of intelligent services in complex industrial settings.
\end{abstract}

% However, current approaches often follow siloed and tightly coupled architectures, hindering scalability and reuse of intelligent functionalities. In this work, we propose a modular and interoperable approach that enhances the integration of AI pipelines into CPS, minimizing configuration efforts and decoupling the responsibilities of DT and AI components. We introduce the concept of Zero Configuration (ZeroConf) AI pipelines, where DTs act as orchestrators for data collection, assessment, storageand intelligent augmentation. Our approach is validated through an application in a MicroFactory scenario, demonstrating the capability to support multiple ML models concurrently and dynamically manage data-driven processes, thus accelerating the deployment of intelligent services in complex industrial environments.

%%%%%%%%%%%%%%%%%%%%%%%%%%%%%%%%%%%%%%%%%%%%%%%%%%%%%%%%%%%%%%%%%%%%%%%%%%%%%%%%
\section{Introduction}
\label{sec:introduction}

The increasing pervasiveness of Cyber-Physical Systems (CPS) in the industrial domain has brought forth unprecedented opportunities for data-driven intelligence, but also significant challenges rooted in the inherent complexity of these environments. Industrial systems are composed of a vast and heterogeneous network of devices and components, often deployed across distributed settings and relying on a fragmented ecosystem of Internet of Things (IoT) and Industrial IoT (IIoT) technologies \cite{cimino_review_2019}. This fragmentation is reflected in the diversity of communication protocols, data formats and device capabilities, which collectively hinder the seamless integration of digital services. As a result, developing general purpose intelligent applications based on Artificial Intelligence (AI) and Machine Learning (ML) with minimized configuration and adaptation requirements remains a challenging task. The difficulty lies not only in discovering and interfacing with the physical world but also in understanding the semantics, quality and context of the data being collected. The impact of key data-quality issues on a typical intelligent pipeline is summarized in Tab. \ref{tab:mlops_pipeline_metrics}. This complexity forms a substantial gap between low-level, device-centric layers and the high-level demands for intelligent, autonomous functionalities. Bridging this gap requires architectural abstractions and integration strategies that can reduce heterogeneity, provide meaningful context and support scalable AI deployments.

In recent years, the concept of the Digital Twin (DT) has been revitalized and increasingly adopted as a promising architectural paradigm to address the complexity of cyber-physical systems, particularly in industrial settings \cite{Minerva, 9640612}. By creating digital, interoperable replicas of physical assets, often referred to as Physical Twins (PTs), DTs provide a structured and semantically rich abstraction layer that facilitates interaction with the physical world. This digital mirroring simplifies the introduction of intelligent functionalities, such as those enabled by ML, by offering consistent, contextualized and high-quality data representations. However, current implementations are often characterized by siloed architectures, where the physical layer, digital representation and intelligent modules are tightly coupled in domain-specific ways. These vertically integrated solutions limit scalability, reuse and interoperability, creating barriers to the widespread deployment of AI-driven functionalities \cite{8220372, LIPPI2023103892}.

This scenario highlights a clear research opportunity: to define a more modular and interoperable approach that enables seamless integration of AI pipelines into cyber-physical systems as initially explored in \cite{LOMBARDO2024107203}. Such an approach would minimize configuration and customization efforts by decoupling responsibilities across components. In this vision, the DT would not only act as a convergence point for physical-digital interaction but also manage data assessment and pre-processing, support storage and lifecycle management for model training and retraining and enable concurrent testing of multiple ML models. Assigning clear responsibilities to DT and AI modules can thus enhance modularity, improve scalability and accelerate the deployment of intelligent services in complex industrial environments.

\begin{comment}
\begin{table*}[ht!]
\centering
\renewcommand{\arraystretch}{1.1}
\setlength{\tabcolsep}{6pt}
\caption{Expected Impact of Digital Twin + ZeroConf on Key Data Quality and AI Project Metrics}
\label{tab:zeroconf_impact}
\resizebox{\textwidth}{!}{%
\begin{tabular}{p{3.5cm} p{2.5cm} p{2.5cm} p{2.5cm} p{4cm}}
\toprule
\textbf{Metric} & \textbf{Baseline} & \textbf{Improvement} & \textbf{Post ZeroConf} & \textbf{Source} \\
\midrule
Time spent on data cleaning 
  & 60--80\% 
  & ↓40 pp 
  & 36--48\% 
  & Forbes (2020); Dataversity (2018) \\[1ex]

AI project failure rate 
  & 85\% 
  & ↓25 pp 
  & 60\% 
  & Gartner (2021) \\[1ex]

Cost of poor data quality 
  & \makecell[l]{\$3.1 T/year (US)\\\$9.7 M/year (per org.)} 
  & ↓30\% 
  & \makecell[l]{\$2.17 T/year (US)\\\$6.8 M/year (per org.)} 
  & IBM (2016); Gartner (2019) \\[1ex]

Bias amplification 
  & 15--25\% 
  & ↓10 pp 
  & 5--15\% 
  & Smith et al.\ (2020) \\
\bottomrule
\end{tabular}}
\vspace{0.1cm}

\noindent
\textbf{Note:} Arrows (↓) indicate absolute percentage-point reductions.  
\end{table*}
\end{comment}

This cyber-physical abstraction allows for the design of ML pipelines that minimize human intervention, adopting a \textit{Zero Configuration} (ZeroConf) approach that reduces initial setup and manual involvement. This approach also supports the automated selection of modules for signal processing and segmentation and data cleaning by leveraging a catalog of general-purpose functions configurable on demand. 

% Automatic signal segmentation is crucial in industrial settings, where identifying a machine’s operational phases is often a prerequisite for applying ML techniques. These phases can be determined either through direct machine data or inferred using advanced signal processing and ML methods, with Digital Twins (DTs) serving as the cyber-physical software bridge that enables efficient data preparation. DTs also offer a powerful infrastructure for running multiple ML models in parallel, allowing the evaluation of alternative approaches and functionalities in complex scenarios. By structuring data and providing the necessary operational context, DTs enable automated ML pipelines and dynamically orchestrate data preparation, segmentationand inference.

This work presents three main contributions: an analysis of the capabilities and properties of DTs that enhance AI/ML integration in industrial systems; the introduction of a ZeroConf AI/ML pipeline concept, outlining phases, modules and functional categories for rapid, configuration-free deployment; and the validation of the proposed approach in a MicroFactory environment, demonstrating the practical impact of DT-driven pipelines applied to accelerometric data.

\begin{table}[ht]
  \centering
  \renewcommand{\arraystretch}{1.1}
  \setlength{\tabcolsep}{4pt}
  \caption{Key data‐quality issues and their quantified impact on a typical MLOps pipeline.}
  \label{tab:mlops_pipeline_metrics}
  \begin{tabular*}{\columnwidth}{@{\extracolsep{\fill}} 
      p{0.50\columnwidth}  % Issue
      p{0.29\columnwidth}  % Impact
      p{0.38\columnwidth}  % Source
    }
    \toprule
    \textbf{Issue}                      & \textbf{Impact}     & \textbf{Source}              \\
    \midrule
    Time spent on data cleaning         & 60–80\%             & \cite{press2016,guess2016}   \\
    AI project failure rate             & 85\%                & \cite{dynatrace2021}         \\
    Cost of poor data quality           & \$3.1 T/yr (US)     & \cite{redman2016}            \\
    Bias amplification in models        & 15–25\%             & \cite{redman2016}             \\
    Model deployment lead time          & 2–3 weeks           & \cite{mckinsey2020}          \\
    Retraining turnaround time          & 3–5 days            & \cite{aws2021}               \\
    Inference latency in production     & 100–200 ms          & \cite{cloud2019}             \\
    CI/CD pipeline failure rate         & 30–40\%             & \cite{devlakeCFR}            \\
    \bottomrule
  \end{tabular*}
\end{table}

%%%
\begin{figure*}[ht]
    \setlength{\belowcaptionskip}{-13pt}
    \centering
    \includegraphics[width=\textwidth]{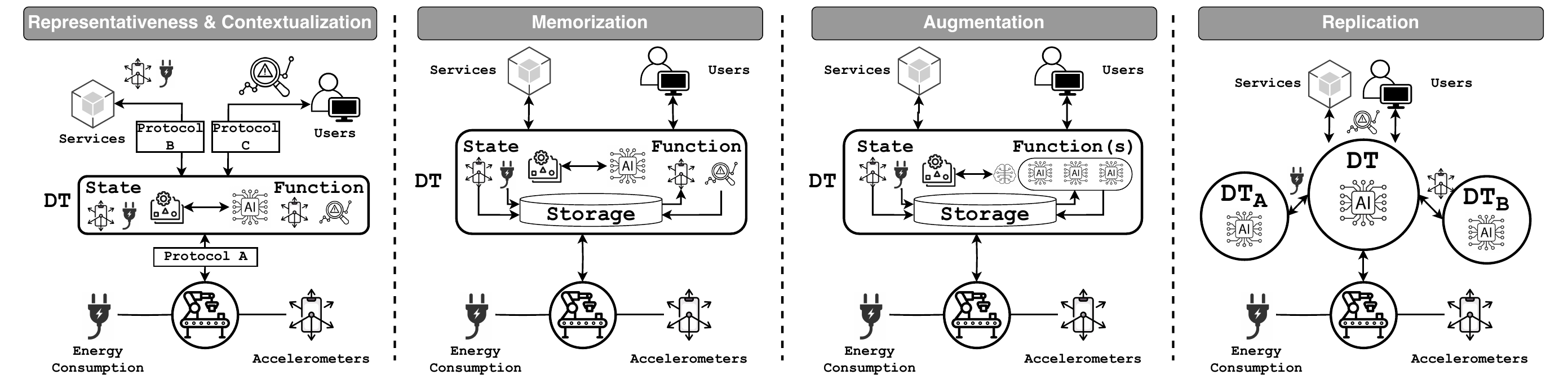}
    \caption{Schematic representation of how core DT capabilities interconnect and support AI functionalities within a DT instance.}
    \label{fig:dt_capabilities_ai_functions}
\end{figure*}
%%%

\section{Related Works}
\label{sec:related}

Cyber-Physical Systems (CPS), systems that integrate computational and physical elements, particularly in industrial domains, are inherently heterogeneous, combining diverse equipment, communication protocols and data modalities. This heterogeneity results in the continuous generation of large volumes of unstructured or semi-structured data, which are increasingly difficult to manage and interpret. As data volumes scale, extracting actionable value becomes not only a competitive advantage but a necessity, driving the integration of AI/ML solutions as essential tools for intelligent system behavior.

Within this context, the concept of the DT has gained traction as a promising abstraction layer to manage and bridge the complexity of physical systems and their digital counterparts. Initially conceptualized by Michael Grieves in the early 2000s in the context of Product Lifecycle Management (PLM) \cite{DT-concept-thread-shadow-PLM-Tao-2019,origins-DT-concept-Grieves-2016}, DTs have since evolved beyond their original scope. Later refinements, such as the classification by Kritzinger et al. \cite{DT-in-manufacturing-review-Kritzinger-2018}, distinguish between digital models (DM), digital shadows (DS) and fully interactive DTs, the latter enabling bidirectional synchronization and real-time decision making. Key enabling technologies for effective DT deployment include IoT/IIoT infrastructures, which provide the necessary real-time sensing, actuation and AI/ML models, which extract insights and enable autonomous system behavior. AI integration within DTs enhances system intelligence by supporting predictive maintenance, anomaly detection, adaptive control and optimization \cite{JAN2023119456}. Forecasting applications, ranging from oil production \cite{forecast-oil-production-Weiss-2002} to spare parts demand \cite{demand-forecast-machinery-company-Aktepe-2021} and production time estimation \cite{big-data-driven-job-remaining-time-prediction-deep-learning-approach-Fang-2020}, demonstrate the growing relevance of AI in industrial CPS.

Despite these advancements, current AI deployments in CPS tend to be highly application-specific and domain-tailored \cite{ai-for-I40-review-Jan-2023}. This fragmentation poses significant challenges to model reuse and generalization, particularly in increasingly dynamic and customized manufacturing contexts \cite{exploring-complexity-discrete-manufacturing-Modrak-2019,complexity-design-manufacturing-Monostori-2012}. As production environments become more complex and less standardized, scalable and adaptable AI integration becomes increasingly difficult.

These limitations are compounded by the absence of standardized mechanisms for managing data quality, model performance evaluation and context-aware integration across industrial systems. As a result, effective AI deployment on MCUs and within broader CPS architectures often requires significant manual configuration, custom integration and tightly coupled software stacks, factors that hinder reusability, scalability and long-term maintainability.

This landscape highlights a clear research opportunity: the development of a ZeroConf approach that can abstract the underlying complexity of CPS. By embedding semantic data descriptions, modular AI execution environments and dynamic orchestration mechanisms into DT, the envisioned ZeroConf strategy could reduce the need for manual intervention, maximize the reusability of AI models across applications and enable more agile, scalable deployment of intelligent functionalities.

\section{DT Capabilities \& ZeroConf Intelligence}
\label{sec:dt_capabilities_zeroconf_intelligence}

DTs have become pivotal in connecting the physical and digital dimensions of cyber-physical systems. Their foundational capabilities such as representativeness, memorization, augmentation and replication (all summarized in the Tab. \ref{fig:dt_capabilities_ai_functions}), enable the structured organization and scalable integration of intelligent functionalities \cite{Minerva}. These properties form the basis for designing \textit{ZeroConf AI pipelines}: autonomous workflows that can ingest, interpret and act on data minimizing manual setup, schema alignment, or data preprocessing. This level of automation can be effectively supported and enabled by the DT’s ability to semantically contextualize both real-time and historical data, enforce data validity and orchestrate embedded AI components in a dynamic and adaptive manner.

\begin{table*}[ht!]
  \centering
  \renewcommand{\arraystretch}{1.1}
  \setlength{\tabcolsep}{6pt}
  \caption{Mapping of Digital Twin capabilities to features and benefits in a ZeroConf pipeline}
  \label{tab:dt_capabilities}
  \resizebox{\textwidth}{!}{%
    \begin{tabular}{p{3cm} p{6.5cm} p{6.5cm}}
      \toprule
      \textbf{DT Capability} & \textbf{Features} & \textbf{Benefits} \\
      \midrule
      \textbf{Representativeness}
        & Automated outlier detection \& removal (e.g.\ $7\sigma$ threshold); gap-filling, smoothing \& normalization; rolling-maximum peak extraction; new semantical representations
        & High-SNR, bias-free inputs; consistent, structured data; zero manual preprocessing \\

      \textbf{Memorization}
        & Persistent time-series archives with versioning; sliding-window queries; contextual tagging (timestamps, hyperparameters); retention of historical system behavior
        & Enables trend analysis, drift detection \& retraining; supports historical forecasting; facilitates temporal feature engineering with minimal manual effort \\

      \textbf{Augmentation}
        & Embedded AI services (anomaly detection, forecasting); dynamic signal enrichment; adaptive thresholding; synthetic scenario generation for what-if analysis
        & Adds intelligent, real-time analytics; supports proactive monitoring \& optimization; enables seamless extension of DT functionalities \\

      \textbf{Replication \& Versioning}
        & Parallel DT replicas with varied hyperparameters; automated silhouette scoring \& comparative benchmarking; ensemble A\slash B testing; enable what-if scenarios
        & Identifies optimal hyperparameters with minimal oversight; guarantees reproducibility; accelerates scalable deployment \\
      \bottomrule
    \end{tabular}%
  }
\end{table*}

% DTs have become essential enablers for bridging the physical and digital worlds in cyber-physical systems. Their core capabilities—representativeness, memorization, augmentationand replication—provide the foundation for structuring, simplifyingand scaling intelligent functionalities. By leveraging these capabilities, it is possible to design ZeroConf AI pipelines that minimize manual intervention and support modular, adaptive deployments. A ZeroConf intelligent pipeline refers to an AI/ML processing workflow that can automatically ingest, processand reason over data from cyber-physical systems without requiring manual configuration, schema alignment, or pre-processing steps. This automation is achieved through the intrinsic properties of Digital Twins, including their ability to contextualize, structureand validate real-time and historical dataand to execute embedded intelligent components dynamically.

In this section, we analyze how DT capabilities (schematically represented also in Figure \ref{fig:dt_capabilities_ai_functions}) can be harnessed to streamline the integration of intelligent functionalities, promoting a more flexible and efficient adoption of AI within cyber-physical environments.

\paragraph{Representativeness \& Contextualization} A key capability of DTs lies in their representativeness and contextualization. DTs are in charge of providing an accurate and structured abstraction of their PT, encapsulating properties, actions, relationships and events in a semantically meaningful format. This structured abstraction ensures that heterogeneous data collected from the physical world is harmonized into a consistent digital representation. By embedding contextual information directly into the digital model, DTs facilitate downstream processes such as data preprocessing, feature extraction and model training. Intelligent algorithms no longer need to grapple with the diversity of raw data sources; instead, they can interact with a unified, context-enriched data layer. This capability is foundational for ZeroConf pipelines, enabling AI modules to retrieve and process relevant data fields automatically without extensive configuration. In practical terms, for instance, a manufacturing robot's DT could expose harmonized operational data, such as temperature, vibration and activity states, allowing anomaly detection or predictive maintenance models to be seamlessly deployed.
Moreover, by embedding rich semantic metadata into each feature vector, DTs unlock not only advanced data preprocessing through an abstraction layer but also advanced \emph{representation learning}\footnote{\textit{Representation learning} automatically discovers useful data features from raw inputs, reducing manual engineering and improving model accuracy, critical for ZeroConf pipelines to work “out-of-the-box.”} (e.g., multimodal sensor fusion), significantly boosting downstream model accuracy while minimizing manual feature-engineering effort.

\paragraph{Memorization} Beyond contextualization, DTs possess memorization capabilities that allow the recording and maintenance of historical data, encompassing changes in states, properties, events and interactions over time. This temporal dimension enriches the digital abstraction, supporting retrospective analyses, trend identification and historical simulations. For AI/ML applications, access to pre-organized historical datasets greatly accelerates the training and validation of models, particularly for tasks like forecasting or anomaly detection. Moreover, memorization enables continuous learning strategies, where models evolve alongside the changing behavior of the physical system. For example, in an industrial production line, a DT logging sensor readings and machine states over months allows ML models to be trained on real-world degradation patterns, enhancing predictive maintenance strategies without requiring complex data extraction procedures.
Furthermore, continuous access to time-stamped histories empowers \emph{continual learning} and automated \emph{concept-drift adaptation}, ensuring models remain calibrated to evolving system behaviors without necessitating full retraining.

%%%
\begin{figure*}[ht]
    \setlength{\belowcaptionskip}{-13pt}
    \centering
    \includegraphics[width=\textwidth]{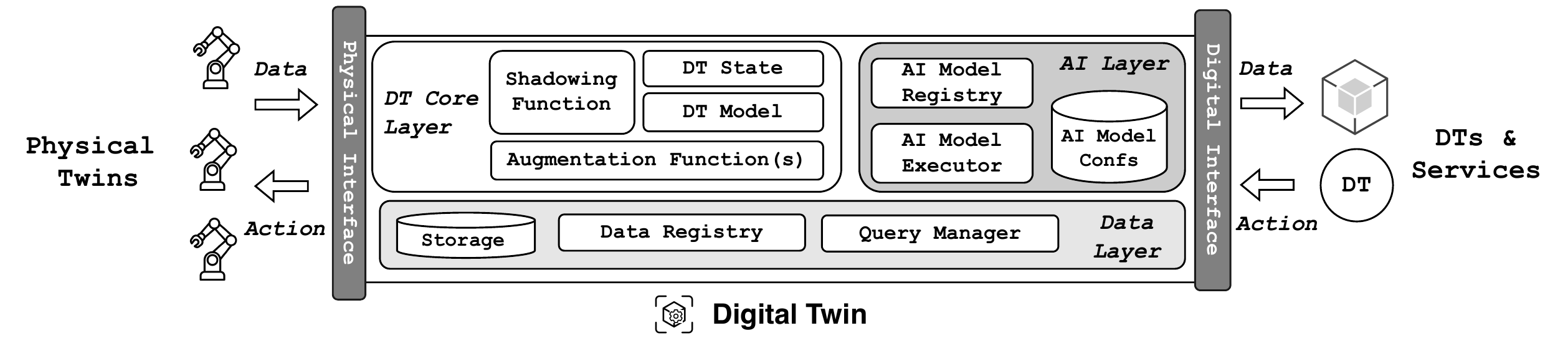}
    \caption{Blueprint architecture of the ZeroConf approach with the three main layers characterizing a DT instance.}
    \label{fig:dt_arch_structure}
\end{figure*}
%%%

\paragraph{Augmentation} This capability further elevates the role of DTs by extending the functionalities of the physical assets through embedded intelligence. Augmented DTs integrate AI/ML models, optimization algorithms and advanced signal processing directly into the digital layer, transforming DTs into hubs of autonomous decision-making and innovation. This dynamic integration supports real-time anomaly detection, operational optimization and predictive capabilities, enabling DTs not only to mirror reality but also to anticipate and act upon it. In an industrial setting, for example, a DT of an electric motor could embed an ML model capable of forecasting bearing wear based on vibration signatures, proactively alerting operators before failures occur. Augmentation is central to ZeroConf pipelines, allowing the flexible addition of intelligent functionalities as modular services that operate autonomously on structured, high-quality data streams.
This capability elevates DTs from passive replicas to active innovation hubs by embedding new, modular services such as synthetic data generators, transfer-learning adapters and what-if simulation engines directly into the digital layer. Augmented DTs can thus repurpose a simple clustering model into a complex forecasting or anomaly-detection pipeline, generate counterfactual datasets on demand and perform domain adaptation across heterogeneous assets. In practice, for example, a motor’s DT might spin off a batch of synthetic failure scenarios, fine-tune an existing vibration-based predictor via transfer learning and run rapid what-if analyses to suggest optimal maintenance schedules. This dynamic augmentation, fully configurable via our configuration, allows intelligence to grow organically on top of the structured, high-quality data streams the DT already provides.

\paragraph{Replication} Replication enables the generation of many DT instances with adjustable configurations from one physical source, each configured for different analytical or operational uses. Ensemble testing, A/B testing and what-if analyses can be conducted by spinning up parallel replicas with alternative hyperparameters, model architectures or data slicing strategies without affecting the live environment. A master replica hierarchy topology cleanly divides tasks, with the master Digital Twin (DT) ensuring a normalized digital model of the asset, while replicas are specialized to specific tasks such as anomaly scoring, throughput prediction, or energy optimization. The output of each replica is versioned and annotated with their configuration metadata, thereby allowing automated benchmarking e.g., silhouette scores, precision/recall and recovery to previous states. In practice a production DT can spawn one copy with a fine grained cluster configuration to capture micro vibrations, another with a coarse model to predict daily output and a third to verify synthetic fault conditions. This replication pattern gives ZeroConf pipelines the ability to scale experiments, choose optimum strategies and roll out updates safely with zero manual reconfiguration at the asset level. Also, containerization renders new capability deployment automatic and modular, eliminating ad hoc configuration from all DT instances. Replicas can be rolled out in shadow mode, executing two versions simultaneously to test performance on live traffic prior to deploying a new release and canary rollouts enable incremental release of the new image to a subset of twins so that regressions can be monitored. Federated updates orchestrate deployments of containers across geographically distributed replicas to ensure consistency and enable fast rollouts. To enhance transparency and trustworthiness, it is feasible to include explainability mechanisms inside each container, providing interpretable outcomes like saliency maps or SHAP values alongside predictions. This comprehensive strategy to replication and versioning ensures ZeroConf pipelines preserve agility, robustness and scalability in changing industrial environments.

Overall, the capabilities of DTs such as structured representation, historical memory, intelligent augmentation and flexible replication, form a robust foundation for enabling scalable, automated AI/ML pipelines in industrial and cyber-physical domains. By leveraging these properties, DTs minimize the complexities associated with system heterogeneity and data fragmentation, dramatically reducing the configuration burden and fostering the realization of ZeroConf intelligent architectures. This convergence of DT technologies and ZeroConf AI/ML pipelines holds the potential to drive the next generation of smart, autonomous industrial systems.
Importantly, replicated DTs support parallel A/B testing of expert models and automated rollback, drastically cutting deployment cycle times and establishing a seamless MLOps feedback loop.

DTs not only encapsulate the real-time state of physical systems but also offer persistent, queryable access to historical data enriched with semantic descriptors. This enables AI components to access relevant input datasets through high-level queries, selecting data based on time windows, machine states, semantic tags, or quality constraints, eliminating the need for hardcoded data extraction pipelines. In parallel, DTs manage \textit{data quality validation}, including freshness checks, missing value detection and semantic consistency verification. These mechanisms ensure the AI models receive reliable and actionable input, increasing overall robustness and allowing self-adaptive behaviors.

\section{DT Blueprint ZeroConf Architecture}
\label{sec:dt_blueprint_arch}

The realization of ZeroConf intelligence through DTs requires a coherent and modular architectural framework. The proposed DT blueprint architecture following and extending state-of-the-art principles and patterns \cite{bellavista2023requirements, web_of_dt}, based on a layered and componentized structure, integrates the core functionalities needed to enable scalable ZeroConf AI pipelines in cyber-physical systems. It comprises distinct yet interconnected layers, namely the \textit{DT Core Layer}, the \textit{Data Layer} and the \textit{AI Layer}, that communicate through well-defined physical and digital interfaces, facilitating structured data flow and service orchestration across the system.

%%%
\begin{figure*}[ht]
    \setlength{\belowcaptionskip}{-13pt}
    \centering
    \includegraphics[width=\textwidth]{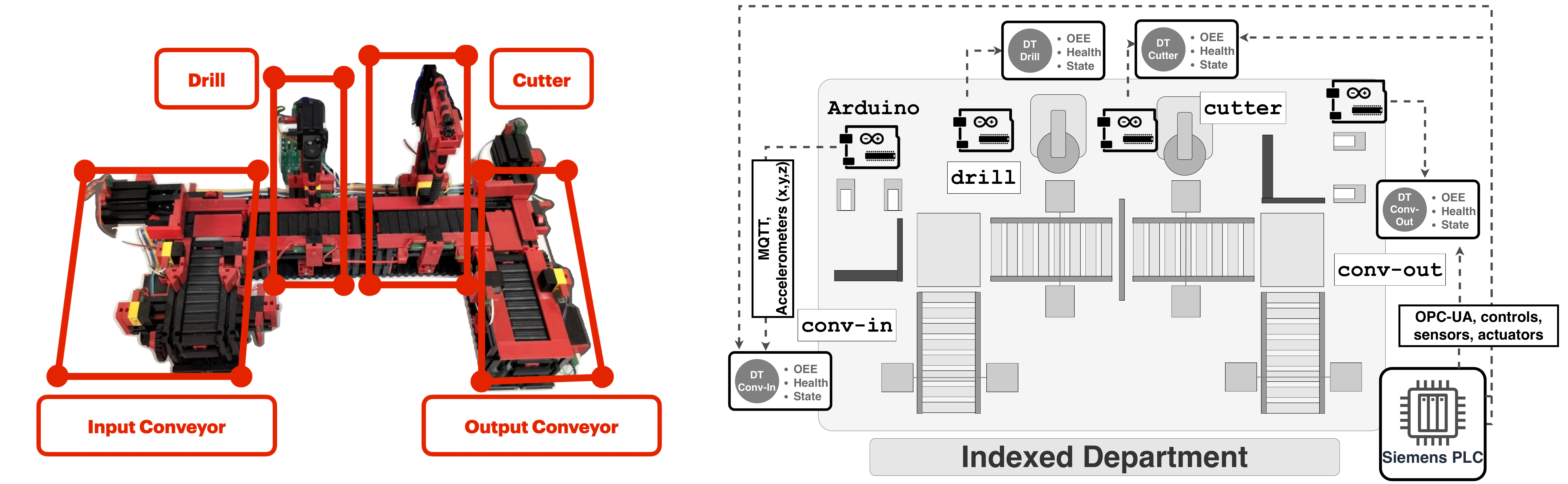}
    \caption{Experimental Microfactory with the different machines, communication protocols, sensors and DTs.}
    \label{fig:mf_dt_schemese-mf-structure}
\end{figure*}
%%%

At the foundation of the architecture (depicted in Figure \ref{fig:dt_arch_structure}) lies the \emph{Physical Interface}, which connects the DT to its corresponding PT. This interface supports bidirectional communication of data and control actions, allowing the DT to monitor real-world behavior and enact decisions in real time. Data ingested through this interface is captured and stored by the \emph{Data Storage} module, which manages both raw sensor data and structured digital representations generated by the DT's \emph{Shadowing Function} \cite{saracco2019dt, web_of_dt}. This function ensures alignment between physical and digital states, enabling synchronous interaction and accurate system modeling. This process computes the current state of the DT, which is structured as a set of: i) \textit{Properties}, representing labeled attributes that evolve in response to changes in the physical asset; ii) \textit{Events}, denoting transient signals derived from real-time observations of the associated physical entity; iii) \textit{Relationships}, reflecting the dynamic connections and dependencies among physical assets and consequently mirrored between their respective DT; and iv) \textit, describing the operations that the DT can expose or trigger on behalf of its physical counterpart, either to send feedback to the physical layer or to leverage services made available through the digital representation\cite{web_of_dt}.

The \emph{Data Layer} combined with the Shadowing Function for the DT's State computation implement core capabilities related to \textit{representativeness} and \textit{memorization}. Automated outlier removal, normalization, smoothing and peak extraction are executed here, ensuring that AI models receive consistent, high-quality inputs without requiring manual preprocessing. Historical data is retained in versioned time-series archives, enriched with contextual metadata (e.g., timestamps, hyperparameters) and made accessible via a \emph{Query Manager}. This enables sliding-window queries and supports advanced AI tasks such as drift detection, temporal analysis and feature engineering. These capabilities provide the foundation for ZeroConf data readiness.

The \emph{DT Core Layer} manages behavioral and structural models of the DT, coordinating state transitions and domain logic. It also plays a key role in contextualizing and labeling data during memorization, enhancing downstream interpretability and reusability for both real-time and historical analysis. This layer ensures consistency in data semantics and supports feature alignment with AI model requirements.

The \emph{AI Layer} builds on this foundation by incorporating the \textit{augmentation} and \textit{replication and versioning} capabilities of the DT. The \emph{AI Model Registry} and \emph{Model Configuration Store} hold model metadata and configuration artifacts necessary for ZeroConf orchestration. Within this layer, the \emph{AI Model Executor} dynamically triggers models using real-time or query-driven inputs and supports embedded analytics functions such as anomaly detection, adaptive thresholds, or clustering extensions. These augmentation features enable seamless integration of advanced analytics directly within the DT context, transforming traditional ML workflows into real-time, embedded services.

To support scalable experimentation and robust deployment, this layer also enables replication and benchmarking. Multiple DT replicas can be instantiated with different hyperparameters, enabling parallel A/B testing, ensemble evaluation and comparative model scoring (e.g., via silhouette metrics). This replication mechanism ensures reproducibility, accelerates model selection and provides a controlled environment for rapid analytics prototyping.

Finally, \emph{Augmentation Functions}, executed either within the AI Layer or coordinated with the DT Core, leverage outputs from trained models to enhance decision-making, automate control strategies, or trigger system adaptations. These functions operationalize the outputs of AI pipelines while maintaining alignment with the DT’s real-time state and data quality expectations.

A DT is a dynamic, continuously evolving software entity. As introduced in~\cite{web_of_dt}, the concept of the DT \emph{lifecycle} defines the distinct operational phases a DT undergoes, from instantiation to termination, focusing on its ability to remain aligned with the state of the corresponding PT. Initially, the DT enters the \texttt{Unbound} phase, during which internal modules are initialized. It transitions to the \texttt{Bound} phase upon successful binding with the PT. When active synchronization is established, the DT enters the \texttt{Synchronized} phase, reflecting a coherent and up-to-date representation of the physical state. Temporary loss of synchronization results in a shift to the \texttt{Out of Sync} phase, which persists until recovery mechanisms reestablish consistency. Finally, when the DT is no longer needed, it progresses to \texttt{Done} and subsequently to \texttt{Stopped}. Faults or reinitializations may cause the instance to loop back to \texttt{Unbound}.

Critically, lifecycle computation plays a strategic role not only in maintaining synchronization but also in enhancing data interpretation and system behavior. Accurate identification and tracking of the current lifecycle phase enable the DT to better segment incoming data streams and correctly associate sensor readings or events with specific operational states of the PT. This alignment is particularly important for applications such as phase-aware anomaly detection or forecasting, where temporal and contextual accuracy is essential. Furthermore, the DT lifecycle is fundamental to supporting ZeroConf operations. By exposing lifecycle metadata, the DT can autonomously decide when to trigger model executions, validate data readiness, or initiate augmentation functions, all without manual intervention. This enables seamless orchestration of intelligent functionalities, even across diverse or dynamic CPS environments. 

Overall, the proposed blueprint establishes a modular, extensible and semantically robust foundation for the deployment of AI-enabled DTs. By clearly delineating functional responsibilities across architectural layers and integrating dynamic data querying, semantic validation and coordinated model execution, the architecture enables the seamless realization of ZeroConf intelligent pipelines. This design not only accelerates the deployment of AI functionalities but also ensures resilience, adaptability and scalability in complex industrial and cyber-physical environments.

\begin{figure*}[htb]
  \centering
  % Prima riga: tre immagini (loss)
  \subfloat{\includegraphics[width=0.32\textwidth]{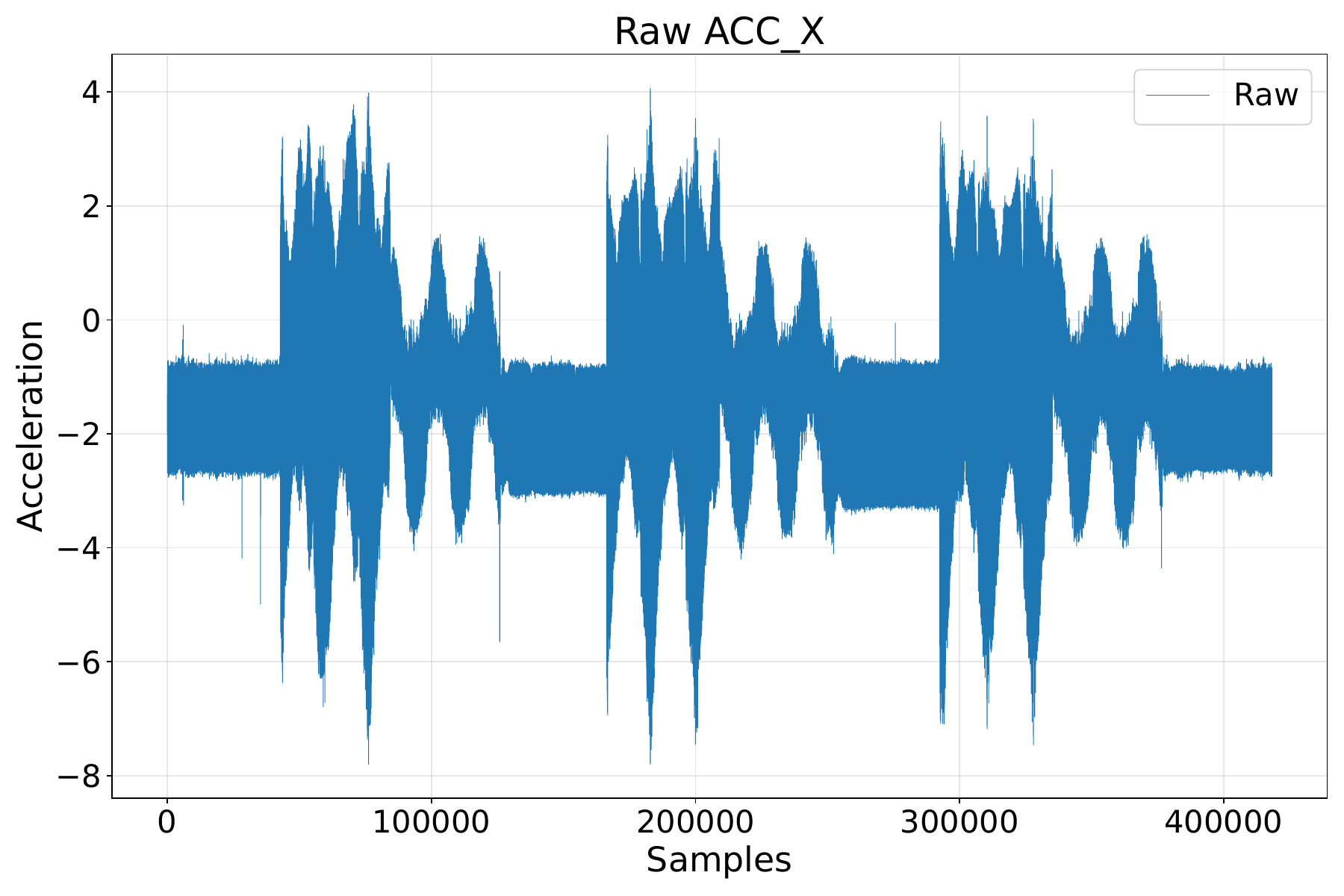}}\hfill
  \subfloat{\includegraphics[width=0.32\textwidth]{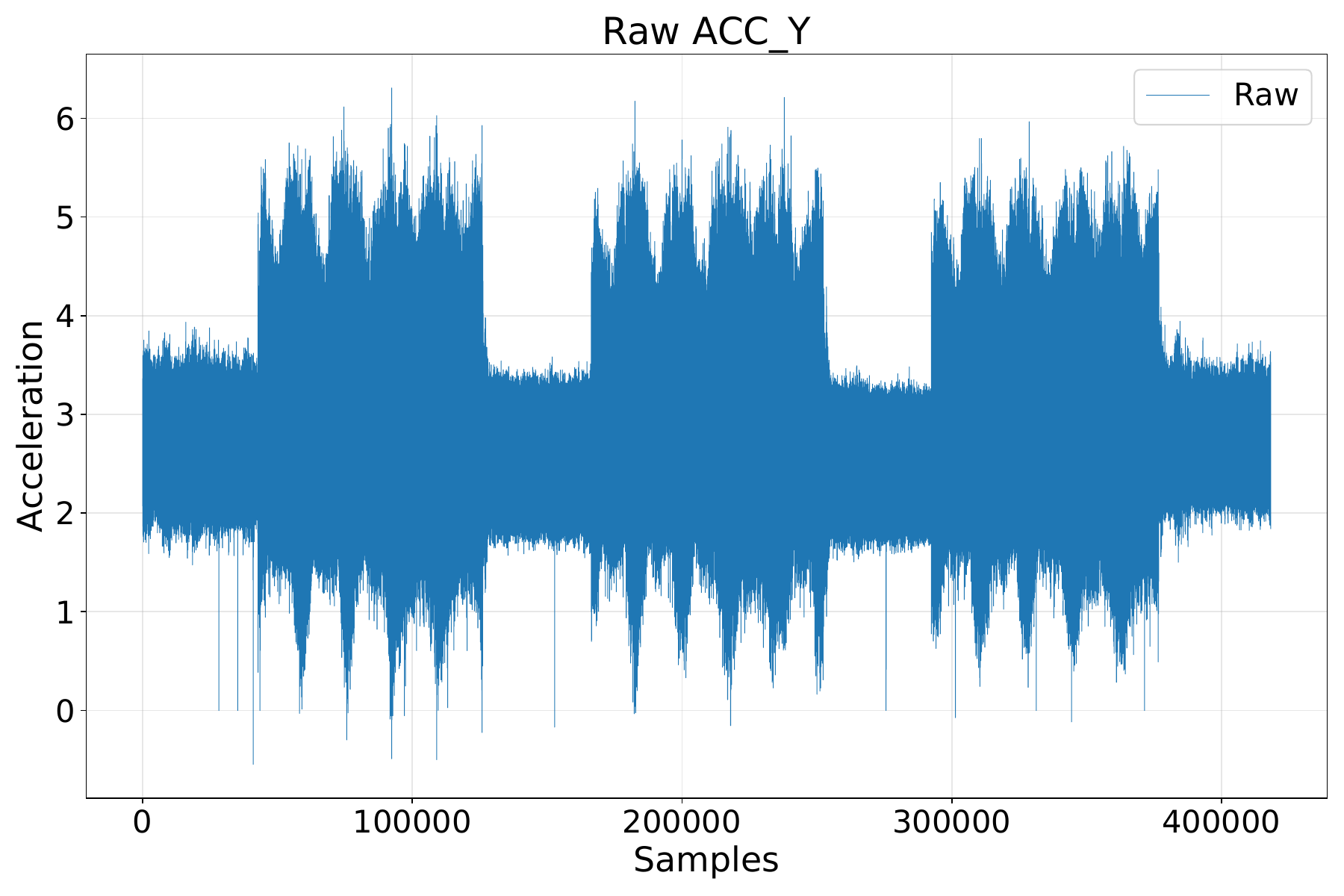}}\hfill
  \subfloat{\includegraphics[width=0.32\textwidth]{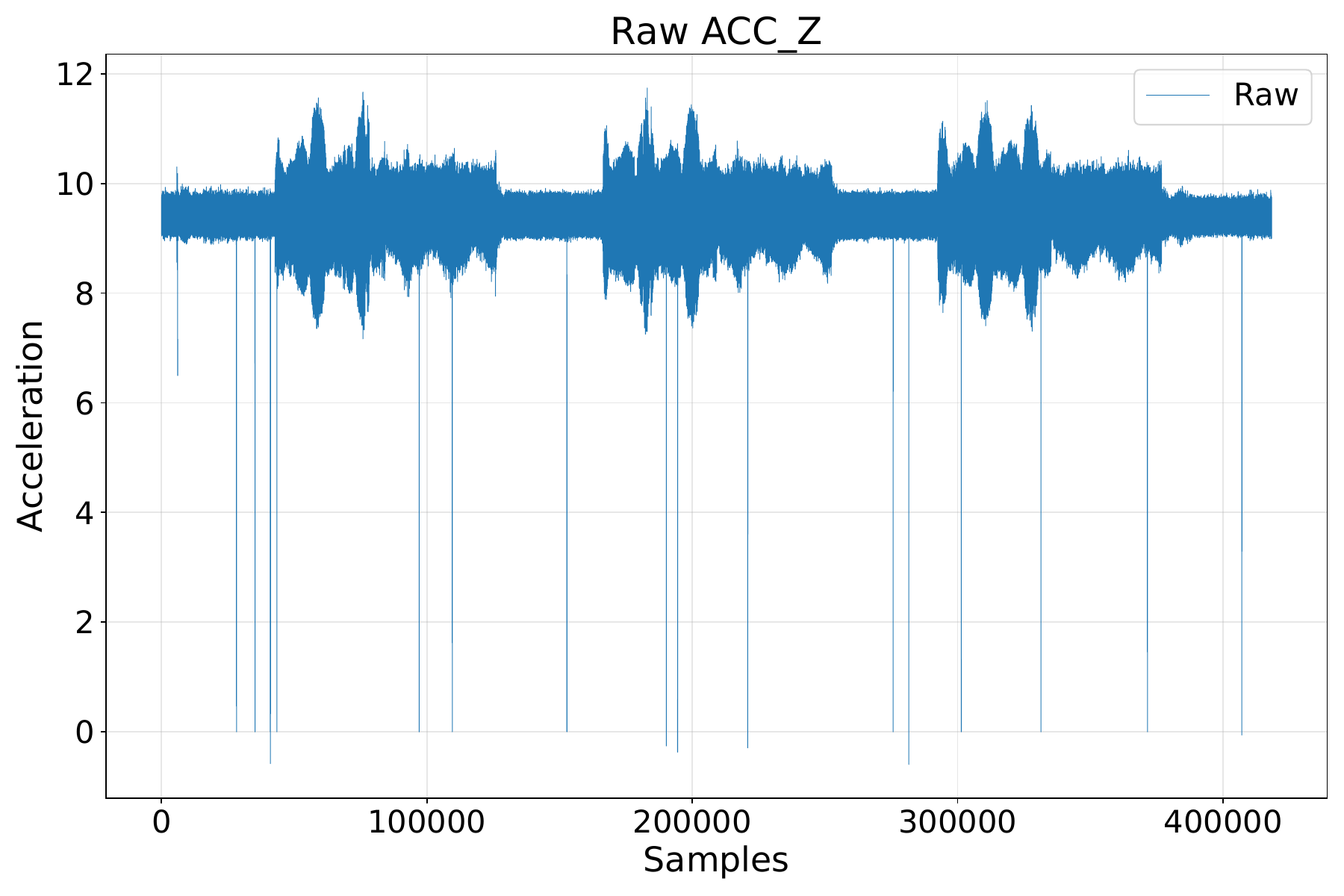}}\\[0.3cm] % spazio verticale ridotto
  % Seconda riga: tre immagini (bar plot)
  \subfloat{\includegraphics[width=0.32\textwidth]{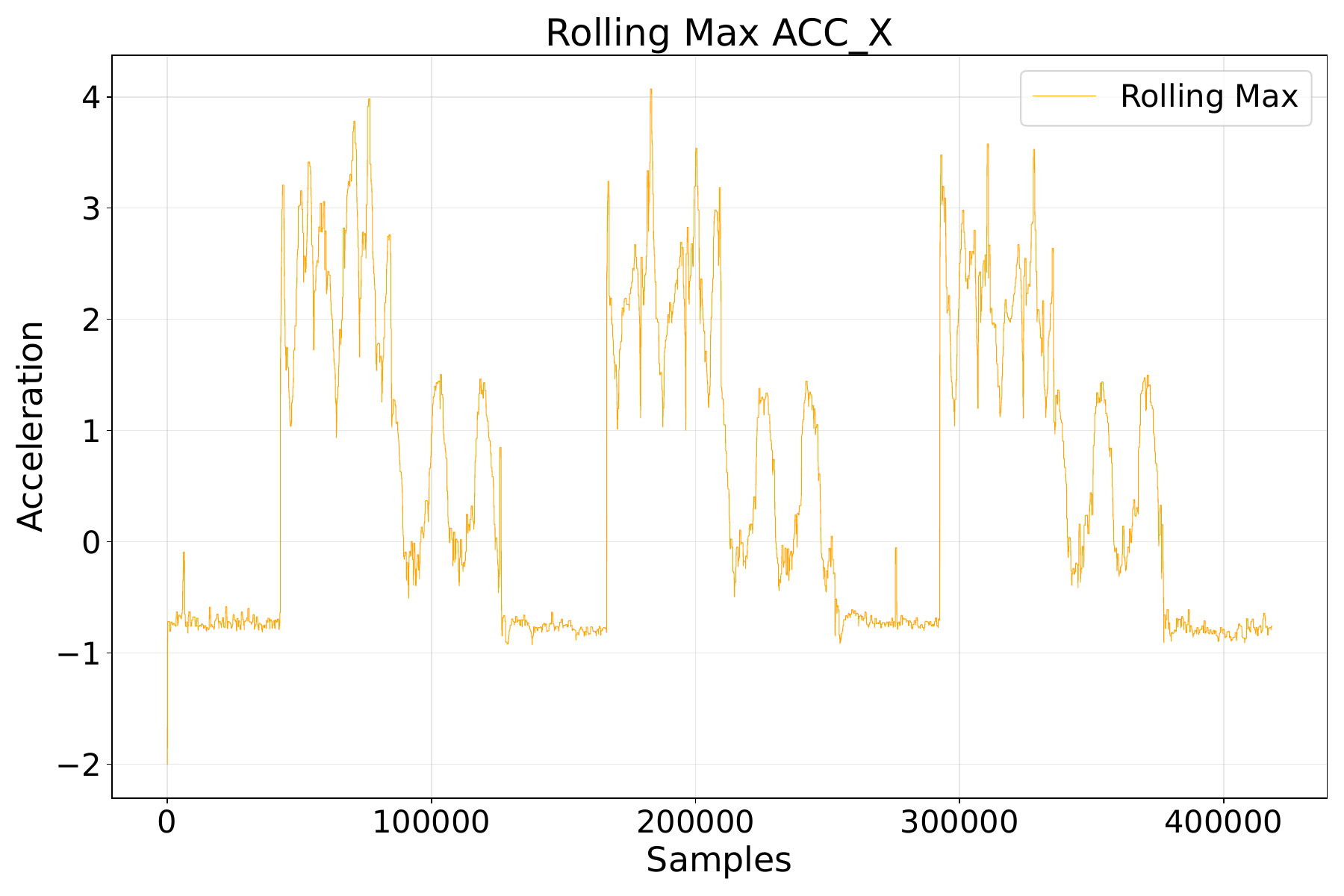}}\hfill
  \subfloat{\includegraphics[width=0.32\textwidth]{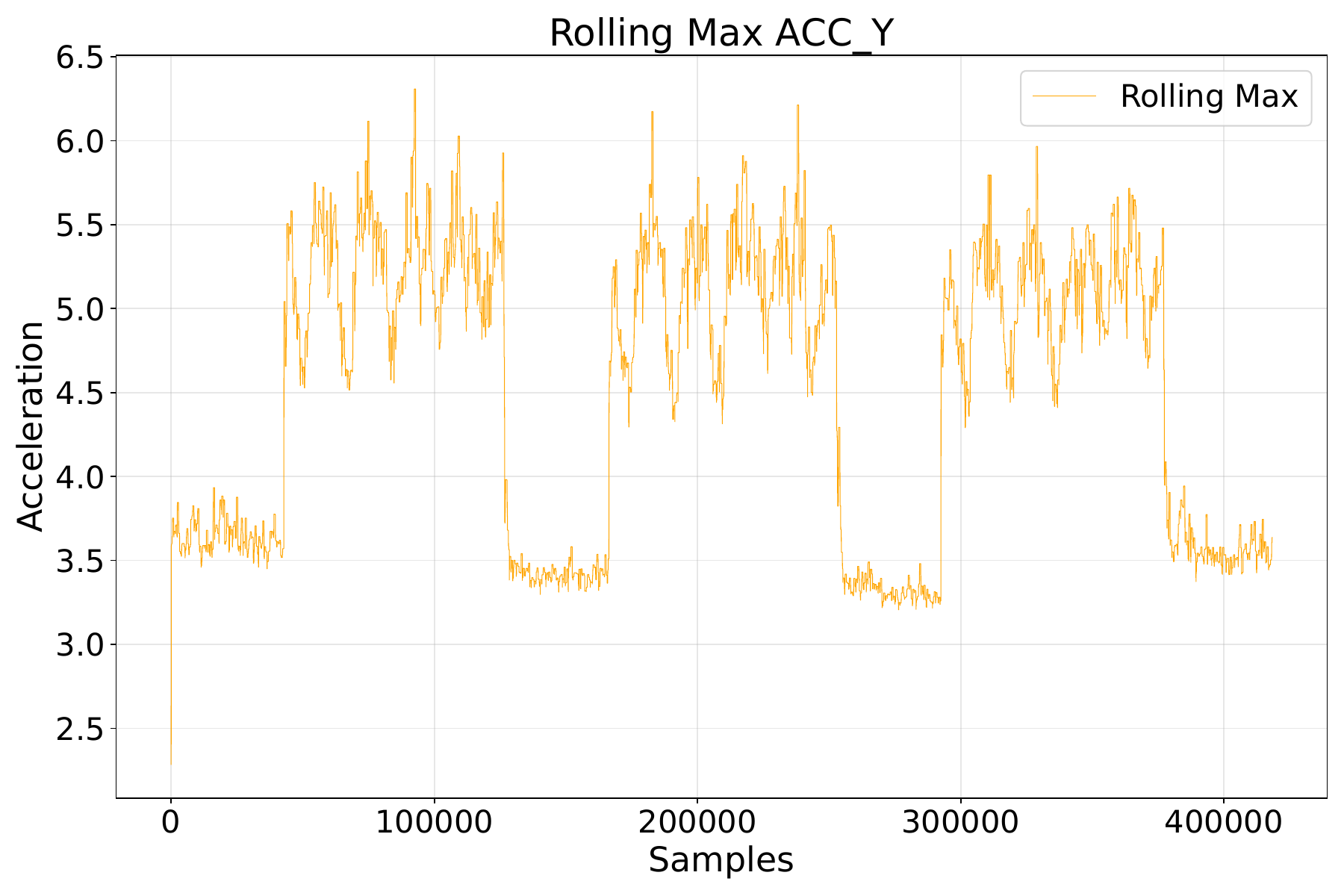}}\hfill
  \subfloat{\includegraphics[width=0.32\textwidth]{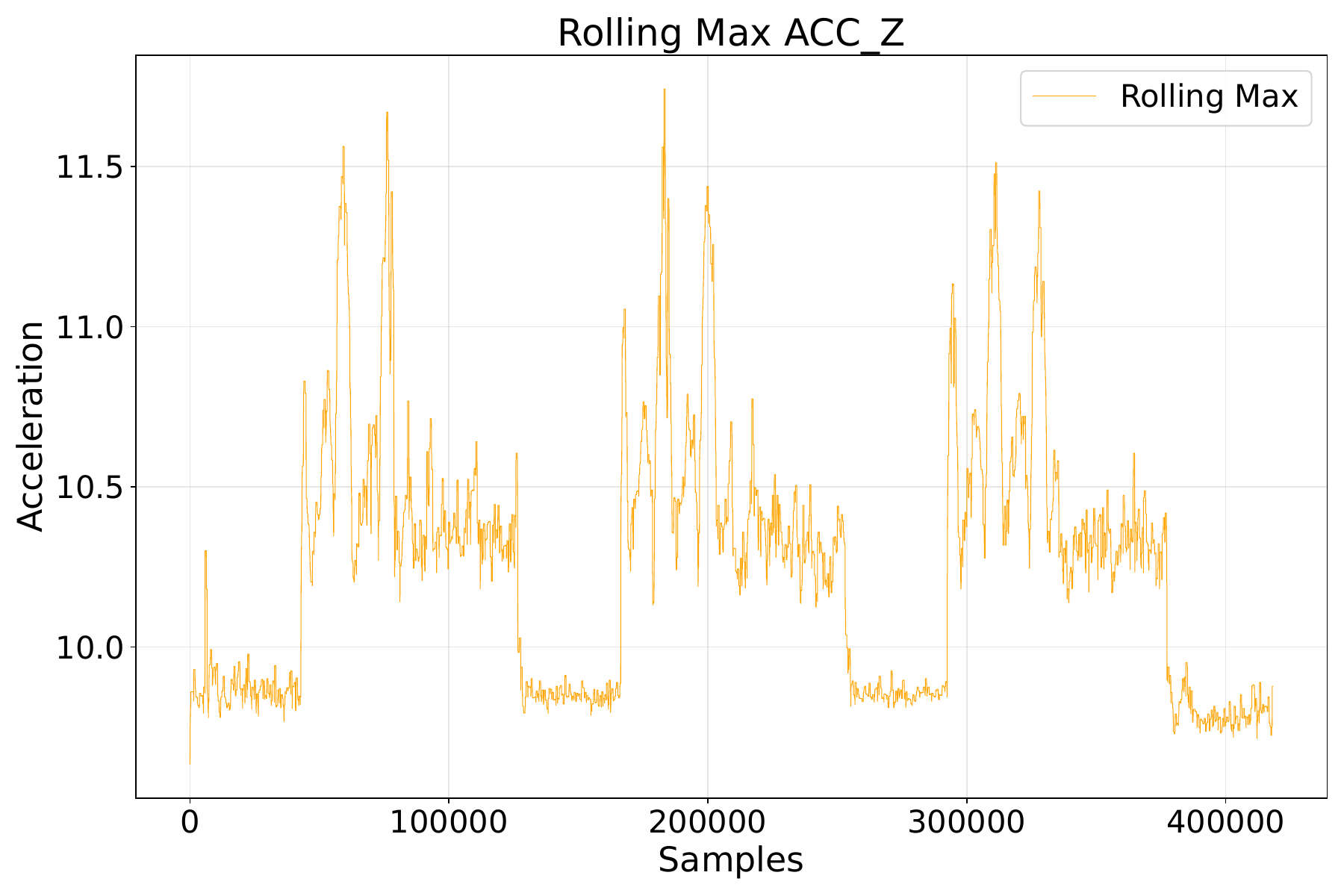}}\\[0.3cm]
  \caption{The \textit{top three charts} show the raw accelerometer signals recorded along the X, Y and Z axes during our microfactory’s experimental phase. Directly beneath each axis plot, we present the corresponding Digital Twin preprocessing step: a rolling‐maximum extraction that condenses each signal into its block‐wise peak values, simplifying the downstream segmentation and clustering.}
  \label{fig:combined}
\end{figure*}

\begin{figure*}[htb]
    \setlength{\belowcaptionskip}{-13pt}
    \centering
    \includegraphics[width=1.0\textwidth]{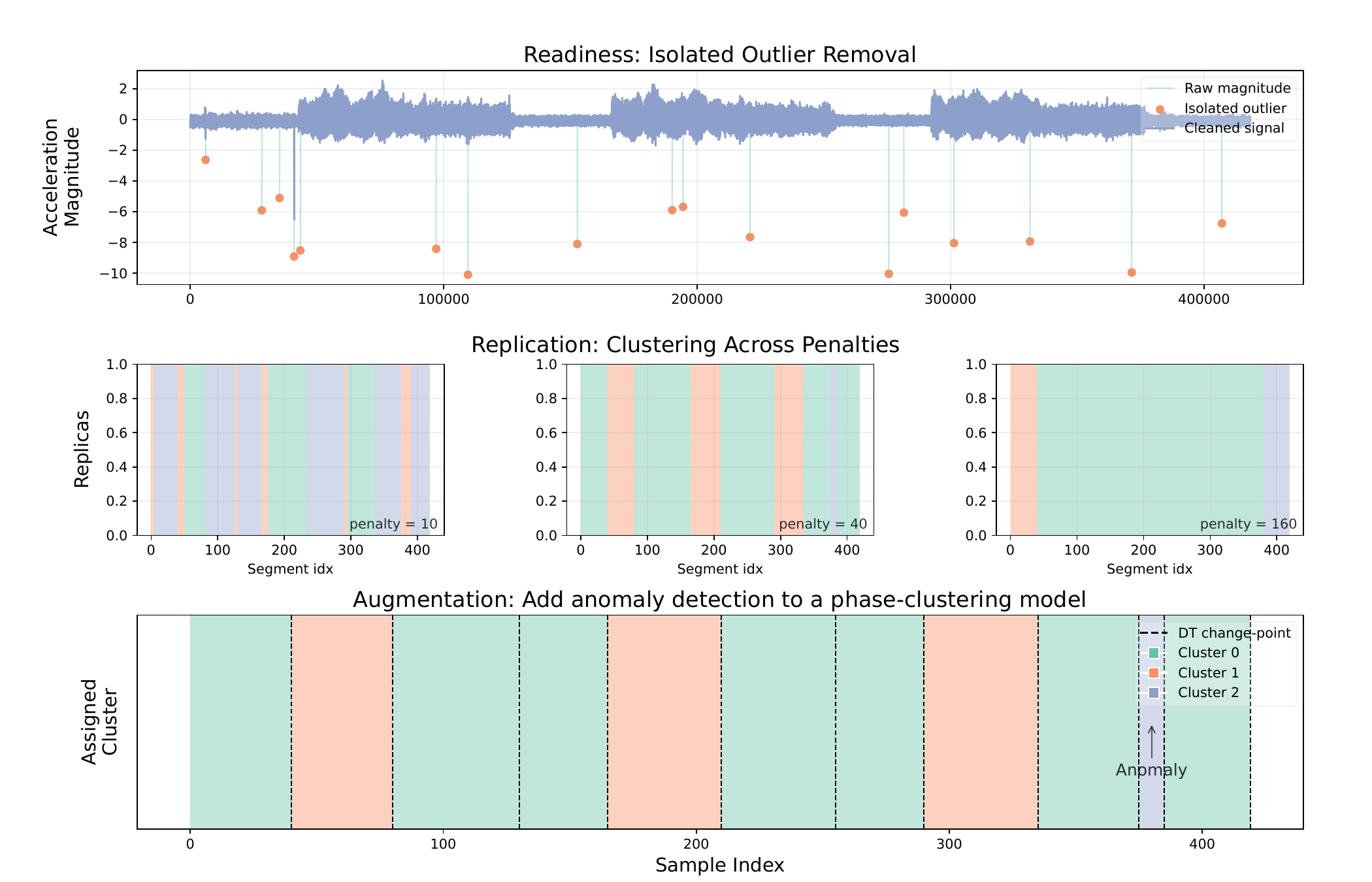}
    \caption{\textit{Top graph} (Readiness): showing cleaned signal through DT. \textit{Middle graph} (Replication): showing panels for penalties with color phases. \textit{Bottom graph} (Augment): showing dashed change-points and colored segments, anomaly marked.}
    \label{fig:core_stages}
\end{figure*}

\section{Experimental Phase Validation}
\label{sec:experimental_evaluation}

% We validate our ZeroConf, DT–based AI pipeline on accelerometer streams from a MicroFactory testing platform, transforming a simple phase clustering model into a complete anomaly detection system by encapsulating it within a ZeroConf-configured Digital Twin. 

We validate our ZeroConf DT pipeline using accelerometer streams collected from a MicroFactory platform, transforming a simple phase-clustering model into a fully automated anomaly detection system embedded within a ZeroConf-configured DT. 

\subsection{Experimental Setup}

The experimental setup is based on the \emph{Fischertechnik Training Factory Industry 4.0}\footnote{Fischertechnik Industry \& Universities: \url{https://www.fischertechnik.de/en/products/industry-and-universities}} Indexed-Line Station, controlled by a Siemens PLC that publishes real-time data via OPC-UA. For vibration monitoring, Arduino RP2040 boards\footnote{Arduino RP2040: \url{https://docs.arduino.cc/hardware/nano-rp2040-connect/}} are employed to capture accelerometer data, which is transmitted using the MQTT protocol\footnote{MQTT Protocol: \url{https://mqtt.org/}}.

The scenario emulates a production line, where the objective of the DT system is to monitor machine operations and evaluate production efficiency. The DT ecosystem architecture, depicted in Figure~\ref{fig:mf_dt_schemese-mf-structure}, comprises four machine-level DTs, each interfacing with the physical equipment via protocol-specific Physical Interfaces (PIs), supporting MQTT and OPC-UA. These DTs continuously monitor machine states (e.g., idle, active, waiting, failure) and extract performance indicators such as Overall Equipment Effectiveness (OEE)\footnote{OEE: \url{https://www.oee.com/}}, calculated from production rate, uptime and downtime.  All DTs are implemented using the open-source WLDT (White Label Digital Twins) framework\footnote{WLDT framework: \url{https://wldt.github.io}}, a Java-based multithreaded platform supporting shadowing, digitalizationand lifecycle management of DTs \cite{wldt_picone_2021}. Within this framework, we developed physical and digital adapters to enable protocol interoperability and device abstraction. 

The designed and developed DTs decouples noisy, biased vibration signals from the analytics core and delivers four complementary capabilities:

\textit{Representativeness \& Contextualization,} the DT ingests raw accelerometer traces X / Y / Z (Fig. \ref{fig:combined}), applies edge detection filters to flag and eliminate isolated spikes exceeding a 7$\sigma$ threshold, performs gap-filling and smoothing through its 'Readiness' function and condenses each block to rolling maximum peaks as shown in Figure , ensuring bias-free inputs for clustering.

\textit{Memorization,} the DT continuously archives segment-level statistics, presenting aggregated cluster frequencies over 24 h alongside overlaid time series windows that reveal evolving trends and periodicities, thereby enabling automated drift detection and on-demand model retraining.

\textit{Augmentation,} the DT injects a change-point detector into the same K-means pipeline, running at approximately 700 Hz to flag rare or unexpected clusters as anomalies and overlay them onto the live signal trace, achieving real-time fault monitoring with high precision and recall.

\textit{Replication \& Versioning,} multiple DT replicas are provisioned in parallel with different hyperparameters such as block size, penalty, cluster count, versioned independently, allowing us to compare silhouette scores and detection outcomes side-by-side and automatically select the optimal configuration without manual code changes.

These results confirm that a ZeroConf-driven DT not only automates data cleaning, model tracking and parameter selection, but also elevates a simple clustering routine into a resilient, low-latency anomaly detection system ideally suited for the unpredictable vibrations of our microfactory environment.

\subsection{Obtained Results}

Fig. \ref{fig:core_stages} illustrates the three core stages of our ZeroConf DT pipeline and the experiments we carried out to validate its necessity. In the upper panel (“Readiness”), the raw X–Y–Z acceleration traces, ordinarily plagued by transient spikes, sensor bias and drift, are automatically filtered: edges beyond a 7 $\sigma$ threshold are removed, missing values interpolated and each block is reduced to its rolling maximum peaks. Without a DT, achieving this level of data hygiene would require bespoke filtering scripts, manual selection of thresholds and offline reprocessing, exposing the clustering engine to erratic inputs and inconsistent segment boundaries. The middle panel (“Replication Across Penalties”) shows side-by-side segmentations for three PELT penalty values (10, 40, 160). Here, the DT’s ZeroConf orchestration spawns multiple replicas of the same pipeline, each with different hyperparameter settings, instantiated by a single DTand versions their outputs in parallel, enabling instantaneous comparison of silhouette scores and segment counts. In a conventional workflow, one would have to clone code, adjust parameters by hand, rerun experiments sequentially and aggregate results manually, a process both error-prone and time-consuming. Finally, the bottom panel (“Augmentation”) overlays real-time change-point detections onto the clustered timeline, flagging rare clusters as anomalies. Integrating such an augmentation layer without a DT would demand custom wrappers around the K-means model, bespoke scheduling code and manual coordination between data-cleaning, segmentation and anomaly routines. The ZeroConf experiment was motivated precisely to avoid these pitfalls: by embedding clustering, preprocessing, replication and anomaly detection inside a self-configuring DT, we achieved a seamless, fully automated pipeline that is robust to noise, free of manual tuning and immediately deployable in our microfactory setting.  

\begin{comment}
\begin{figure*}[htb]
  \centering
  % Prima riga: tre immagini (loss)
  \subfloat{\includegraphics[width=1.0\textwidth]{Figures/three-1.pdf}}\\[0.3cm] % spazio verticale ridotto
  % Seconda riga: tre immagini (bar plot)
  \subfloat{\includegraphics[width=1.0\textwidth]{Figures/three-2.pdf}}\\[0.3cm]
  \caption{Top image: Data Readiness the raw acceleration magnitude.}
  \label{fig:combined}
\end{figure*}
\end{comment}

%\addtolength{\textheight}{-12cm}   

% This command serves to balance the column lengths
% on the last page of the document manually. It shortens
% the textheight of the last page by a suitable amount.
% This command does not take effect until the next page
% so it should come on the page before the last. Make
% sure that you do not shorten the textheight too much.

%%%%%%%%%%%%%%%%%%%%%%%%%%%%%%%%%%%%%%%%%%%%%%%%%%%%%%%%%%%%%%%%%%%%%%%%%%%%%%%%

%%%%%%%%%%%%%%%%%%%%%%%%%%%%%%%%%%%%%%%%%%%%%%%%%%%%%%%%%%%%%%%%%%%%%%%%%%%%%%%%

\section{Conclusions}
\label{sec:conclusions}

This paper has presented an initial step toward the realization of ZeroConf AI pipelines in industrial cyber-physical systems through the use of DTs. By outlining a modular architecture and identifying key capabilities such as data contextualization, historical traceability and embedded AI execution, we proposed a vision where intelligent functionalities could be deployed with minimal manual configuration. The concept has been preliminarily validated in a MicroFactory scenario, demonstrating the feasibility of using DTs to handle data preparation and ML workflows in a streamlined, automated manner. While this work served as a first contribution toward a broader vision, further research was needed to extend and generalize the approach, particularly across diverse domains and more complex AI use cases. Nonetheless, the architecture and design patterns introduced here provided a promising direction for reducing integration complexity and accelerating AI adoption in dynamic industrial environments.

\section*{Acknowledgement}
This work is supported by the European Union - NextGenerationEU - under PRIN 2022, Project ID: 20223N7WCJ and CUP: E53D23007770001, Project Title: TWINKLE - digital TWIN continuum: a Key enabler for pervasive cyber-physicaL Environments.								

\begin{comment}

\end{comment}

\bibliographystyle{IEEEtran}
\bibliography{bib/refs}

\end{document}